\documentclass{article}
\usepackage{spconf,amsmath,graphicx}
\usepackage{algorithm}
\usepackage{algorithmic}
\usepackage{listings}
\usepackage{amsmath}
\usepackage{hyperref}
\usepackage{amssymb}
\usepackage{subcaption}
\usepackage[font=small,belowskip=-15pt,aboveskip=10pt]{caption}

\title{DCFNET: DISCRIMINANT CORRELATION FILTERS NETWORK FOR VISUAL TRACKING}
\name{Qiang Wang, Jin Gao, Junliang Xing, Mengdan Zhang, Weiming Hu}
\address{CAS Center for Excellence in Brain Science and Intelligence Technology,\\ National Laboratory of Pattern Recognition, Institute of Automation,\\ Chinese Academy of Sciences; University of Chinese Academy of Sciences}

\begin{document}
%
\maketitle

\begin{abstract}
Discriminant Correlation Filters (DCF) based methods now become a kind of dominant approach to online object tracking. The features used in these methods, however, are either based on hand-crafted features like HoGs, or convolutional features trained independently from other tasks like image classification. In this work, we present an end-to-end lightweight network architecture, namely DCFNet, to learn the convolutional features and perform the correlation tracking process simultaneously. Specifically, we treat DCF as a special correlation filter layer added in a Siamese network, and carefully derive the backpropagation through it by defining the network output as the probability heatmap of object location. Since the derivation is still carried out in Fourier frequency domain, the efficiency property of DCF is preserved. This enables our tracker to run at more than 60 FPS during test time, while achieving a significant accuracy gain compared with KCF using HoGs. Extensive evaluations on OTB-2013, OTB-2015, and VOT2015 benchmarks demonstrate that the proposed DCFNet tracker is competitive with several state-of-the-art trackers, while being more compact and much faster.
\end{abstract}
\vspace{-0.2cm}
\section{Introduction}
\vspace{-0.1cm}
\label{sec:intro}
Object tracking is a fundamental problem in computer vision with wide applications like human computer interaction and assistant driving systems. One common setting for this problem is to initialize the object of interest in the first frame with a bounding box and the aim is to estimate the trajectory of the object in subsequent frames \cite{OTB2013, OTB100, VOT2015}. Without knowing the target category a priori, tracking of arbitrary objects needs to learn the discriminant information online to achieve high performance. Despite being successfully solved in a tracking-by-detection paradigm \cite{TLD, Struck}, it still remains a challenging problem due to factors like object deformations, appearance variations and severe occlusions. Maintaining real-time speed is also vital for visual tracking, which is usually a bottleneck for many state-of-the-art trackers with online classifiers trained. Recently, discriminant correlation filters (DCF) based trackers \cite{MOSSE, KCF, DSST} achieve an ideal trade-off between accuracy and
\begin{figure}[htb]
\centering
\centerline{\includegraphics[trim=3cm 5.5cm 0cm 4cm,clip,width=8.5cm]{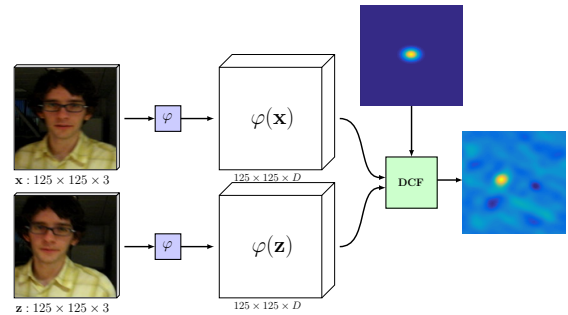}}
\vspace{0cm}
\caption{The overall DCFNet architecture.}
\label{fig:architecture}
\end{figure}\\
speed by solving a ridge regression problem efficiently in Fourier frequency domain. 

Tracking with DCF has been one of the biggest breakthroughs in the community since the hand-crafted multi-channel features (e.g., HoGs \cite{KCF}) were exploited. Trackers going in this direction are always equipped with feature extractors and correlation filters separately. It proves that good features can greatly enhance the tracking performance \cite{wang2015understanding}. Later, more and more works \cite{CF2, DeepSRDCF, HDT} concentrate on integration of multi-layer deep features for DCF tracking. Despite the improved tracking performance, these convolutional layers are usually chosen from the pre-trained networks for the image classification task \cite{AlexNet, VGG, ResNet} or the object detection task \cite{RCNN, SSD}, which are not only hand-picked, but also heavyweight. Since the adopted features in the aforementioned methods are all extracted independently with the correlation tracking process, the achieved tracking performance may not be optimal.

In this paper, we revisit the feature extraction for DCF based trackers. Different from the common DCF approaches which employ existing features, we dissect the closed-form solution of DCF, and find it is natural to develop a network to automatically learn the features best fitting DCF tracking in an end-to-end fashion without hand-interfering. This is surprisingly achieved by treating DCF as a special correlation filter layer added in a Siamese network and carefully deriving the back propagation through it. The architecture of the proposed network (see Fig. \ref{fig:architecture}) contains some convolutional layers which encode the prior tracking knowledge in the off-line training process and constitute a tailored feature extractor. Behind these convolutional layers is the correlation filter layer which can efficiently complete the online learning and tracking by defining the network output as the probability heatmap of object location. To reduce the computational cost, we just set the convolutional layers lightweight. Since the derivation of the correlation filter layer is still carried out in Fourier frequency domain, the efficiency property of DCF is preserved. This enables our tracker to run at high speed of more than 60 FPS during test time, while still achieving competitive tracking accuracy with several state-of-the-art heavyweight and slow trackers.

\vspace{-0.8cm}
\section{RELATED WORK}
\vspace{-0.1cm}
\label{sec:relatedwork}
\textbf{Feature representation for DCF tracking.} The developments on DCF tracking are encouraged continuously over an extended period. MOSSE \cite{MOSSE} first introduces DCF to visual tracking, which runs at high speed of more than 600 FPS simply using the single-channel gray features. CSK \cite{CSK} and KCF \cite{KCF} are the subsequent algorithms which use circulation matrices to interpret correlation filters and generalize to multi-channel feature case. CN \cite{CN} incorporates color names features to boost the performance of CSK. Later, more and more works \cite{CF2, DeepSRDCF} concentrate on integration of pre-trained multi-layer deep features for DCF tracking. HCF \cite{CF2} learns correlation filters on each hierarchical convolutional layer for tracking. DeepSRDCF \cite{DeepSRDCF} builds on only the first layer of single-resolution deep feature maps. Different from the above methods which use the hand-interfering features, we aim to learn the features best fitting DCF tracking automatically. In addition, our network for feature learning is lightweight.\\
\textbf{Other DCF based trackers.} Some works are dedicated to addressing the inherent limitations of DCF tracking. MUSTer \cite{MUSTer} and LCT \cite{LCT} add re-detection mechanisms to achieve long-term DCF tracking. Staple \cite{Staple} incorporates color statistics based model to achieve complementary traits for DCF tracking. DSST \cite{DSST} adds one more scale regression to achieve accurate scale estimation. SRDCF \cite{SRDCF} adds a spatial regularization term to penalize filter coefficients near the template boundaries. Different from them, we aim to bridge the gap between feature extractors and correlation filters.\\
\textbf{Other CNN based trackers.} The progress in deep learning spreads to tracking field remarkably \cite{DLT, MDNet, FCNT, CF2, SiameseFC}. Some works \cite{DLT, FCNT, MDNet} follow the off-line training and online fine-tuning paradigm, which is somehow time consuming for real-time tracking. The correlation filter layer in our network also needs to be updated online. However, since the derivation of it is carried out in Fourier frequency domain, the efficiency property of DCF is preserved. Some works \cite{SiameseFC, GOTURN} also use the Siamese network to build template matching based trackers without online updating, and achieve high tracking speed. Different from them, our network can be incrementally updated, and thus can be regarded as a RNN network in this spirit (see Sec. \ref{subsec:rnn}).

\vspace{-0.1cm}
\section{The Proposed Network}
\vspace{-0.2cm}
\label{sec:dcfnet}
In this section, we first introduce the preliminaries of discriminant correlation filters. Second, we detail the derivation of the backpropagation. Finally, we introduce the online tracking process and give an explanation in the spirit of RNN.
\vspace{-0.2cm}
\subsection{Discriminant correlation filters}
\vspace{-0.1cm}
In the standard discriminant correlation filters, we train a discriminative regression on the features of target patch $\varphi(\mathbf{x}) \in \mathbb{R}^{M\times N\times D}$ and the ideal response $\mathbf{y}\in \mathbb{R}^{M\times N}$ which is a gaussian function peaked at the center. The desired filter $\mathbf{w}$ can be obtained by minimizing the output ridge loss:
{\setlength\abovedisplayskip{1pt}
\setlength\belowdisplayskip{1pt}
\begin{equation}
\label{eq:problem}
\epsilon =
\left \| \sum_{l=1}^{D} \mathbf{w}^{l}\star \varphi^{l}(\mathbf{x})-\mathbf{y} \right \|^{2} +\lambda \sum_{l=1}^{D}\left\| \mathbf{w}^{l} \right\|^{2}
\end{equation}
where $\mathbf{w}^{l}$ refers to the channel $l$ of filter $\mathbf{w}$, $\star$ means circular correlation and the constant $\lambda \geq 0$ is regularization coefficient. The solution can be gained as \cite{DSST}:
\begin{equation}
\label{eq:wsolution}
\hat{\mathbf{w}}^{l}=
\frac{\hat{\varphi}^{l}(\mathbf{x}) \odot \hat{\mathbf{y}}^*}
{\sum_{k=1}^{D}\hat{\varphi}^{l}(\mathbf{x}) \odot (\hat{\varphi}^{l}(\mathbf{x}))^* +\lambda}
\end{equation}
Here, the hat $\hat{\mathbf{y}}$ denotes the discrete Fourier transform $\mathcal{F}(\mathbf{y})$, $y^{*}$ represents the complex conjugate of a complex number $y$ and $\odot$ denotes Hadamard product.

For the detection process, we crop a search patch and obtain the features $\varphi(\mathbf{z})$ in the new frame, the translation can be estimated by searching the maximum value of correlation response map $\mathbf{g}$ , see \cite{DSST} for more details:
\begin{equation}
\label{eq:detection}
\mathbf{g}=\mathcal{F}^{-1}\left( \sum_{l=1}^{D} \hat{\mathbf{w}}^{l*} \odot \hat{\varphi}^l(\mathbf{z}) \right)
\end{equation}
\vspace{-0.5cm}
\subsection{DCFNet derivation: backpropagation}
\vspace{-0.1cm}
The traditional DCF based trackers can only tune the hyper-parameters heuristically, while we can tune them and the feature extraction parameters simultaneously. As shown in Fig. \ref{fig:architecture}, the network is realized by cascading a feature extractor with a DCF modul to get the response of object location. Giving the features of search patch $\varphi(\mathbf{z})$, the desired response $\tilde{\mathbf{g}}$ should get a high response at the real location. The objective function can be formulated as:
\begin{equation}
\label{eq:objective}
\begin{aligned}
L(\boldsymbol{\theta}) &= \left \| \mathbf{g} - \tilde{\mathbf{g}} \right \|^{2} +\gamma \left\| \boldsymbol{\theta} \right\|^{2} \\
\textrm{s.t.} \quad\quad
\mathbf{g} &= \mathcal{F}^{-1}\left( \sum_{l=1}^{D} \hat{\mathbf{w}}^{l*} \odot \hat{\varphi}^l(\mathbf{z},\boldsymbol{\theta}) \right)\\
\hat{\mathbf{w}}^{l}  &=
\frac{\hat{\mathbf{y}}^* \odot \hat{\varphi}^l(\mathbf{x},\boldsymbol{\theta})}
{\sum_{k=1}^{D}\hat{\varphi}^{k}(\mathbf{x},\boldsymbol{\theta}) \odot (\hat{\varphi}^{k}(\mathbf{x},\boldsymbol{\theta}))^*+\lambda}
\end{aligned}
\end{equation}
A explicit regularization should be incorporated, otherwise, the objective will get a non-convergence condition. We use the weight decay method in the conventional parameter optimization to implicit this regularization. Besides, to restrict the magnitude of feature map values and increase the stability in the training process, we add an LRN layer at the end of the convolutional layers.

Now, let's derive the backward formulas. For simplicity, we start with $\frac{\partial L}{\partial \mathbf{g}}$. The chain rule is a little complicated since the intermediate variables are complex-valued variables. According to \cite{complexbp}, the gradient of discrete Fourier transform and inverse discrete Fourier transform are formulated as:
\begin{equation}
\hat{\mathbf{g}}=\mathcal{F}(\mathbf{g}),\frac{\partial L}{\partial \hat{\mathbf{g}}^*} = \mathcal{F}\left(\frac{\partial L}{\partial \mathbf{g}}\right),
\frac{\partial L}{\partial \mathbf{g}} = \mathcal{F}^{-1}\left(\frac{\partial L}{\partial \hat{\mathbf{g}}^*}\right)
\end{equation}

Since the operations in the forward pass only contain Hadamard product and division, we can calculate derivative per-element:
\begin{equation}
\frac{\partial L}{\partial \hat{\mathbf{g}}_{uv}^*}
=\left( \mathcal{F} \left( \frac{\partial L}{\partial \mathbf{g}}\right)\right)_{uv}
\vspace{-0.1cm}
\end{equation}

For the backpropagation of detection branch,
\begin{equation}
\begin{aligned}
\frac{\partial L}{\partial (\hat{\varphi}_{uv}^l({\mathbf{z}}))^*}
=
\frac{\partial L}{\partial \hat{\mathbf{g}}_{uv}^*}
\frac{\partial \hat{\mathbf{g}}_{uv}^*}{\partial (\hat{\varphi}_{uv}^l({\mathbf{z}}))^*}
=
\frac{\partial L}{\partial \hat{\mathbf{g}}_{uv}^*}
(\hat{\mathbf{w}}_{uv}^l)
\end{aligned}
\end{equation}
\begin{equation}
\begin{aligned}
\frac{\partial L}{\partial \varphi^l({\mathbf{z}})}
= \mathcal{F}^{-1} \left(
\frac{\partial L}{\partial (\hat{\varphi}^l({\mathbf{z}}))^*}
\right)
\end{aligned}
\vspace{-0.1cm}
\end{equation}

For the backpropagation of learning branch, we treat $\hat{\varphi}_{uv}^l({\mathbf{x}})$ and $(\hat{\varphi}_{uv}^{l}({\mathbf{x}}))^*$ as independent variable.
\begin{equation}
\begin{aligned}
\frac{\partial L}{\partial \hat{\varphi}_{uv}^l({\mathbf{x}})}
&=
\frac{\partial L}{\partial \hat{\mathbf{g}}_{uv}^*}
\frac{(\hat{\varphi}_{uv}^l({\mathbf{z}}))^*\hat{\mathbf{y}}_{uv}^*-\hat{\mathbf{g}}_{uv}^*(\hat{\varphi}_{uv}^{l}(\mathbf{x}))^*}
{\sum_{k=1}^{D}\hat{\varphi}_{uv}^{k}(\mathbf{x})(\hat{\varphi}_{uv}^{k}(\mathbf{x}))^*+\lambda}
\end{aligned}
\end{equation}
\begin{equation}
\begin{aligned}
\frac{\partial L}{\partial (\hat{\varphi}_{uv}^l({\mathbf{x}}))^*}
&=
\frac{\partial L}{\partial \hat{\mathbf{g}}_{uv}^*}
\frac{-\hat{\mathbf{g}}_{uv}^*\hat{\varphi}_{uv}^{l}(\mathbf{x})}
{\sum_{k=1}^{D}\hat{\varphi}_{uv}^{k}(\mathbf{x})(\hat{\varphi}_{uv}^{k}(\mathbf{x}))^*+\lambda}
\end{aligned}
\end{equation}
\begin{equation}
\begin{aligned}
\frac{\partial L}{\partial \varphi^l({\mathbf{x}})}
= \mathcal{F}^{-1} \left(
\frac{\partial L}{\partial (\hat{\varphi}^l({\mathbf{x}}))^*}+
\left(
\frac{\partial L}{\partial \hat{\varphi}^l({\mathbf{x}})}
\right)^*
\right)
\end{aligned}
\end{equation}

Once the error is propagated backwards to the real-value feature maps, the rest of the backpropagation can be conducted as traditional CNN optimization. Since all operations of the backpropagation in correlation filter layer are still Hadamard operations in Fourier frequency domain, we can retain the efficiency property of DCF and apply the offline training on large-scale datasets. After the offline training has been completed, we get a tailored feature extractor for online DCF tracking.
\vspace{-0.2cm}
\subsection{Online model update: RNN explanation}
\vspace{-0.1cm}
\label{subsec:rnn}
During the online tracking, we just update the filters $\mathbf{w}$ over time. The optimization problem in Eq. (\ref{eq:problem}) can be formulated in a incremental mode as \cite{CN}.
\begin{equation}
\label{eq:rnnproblem}
\epsilon =  \sum_{t=1}^{p} \beta _t \left( \left \| \sum_{l=1}^{D} \mathbf{w}_p^{l}\star \varphi^{l}(\mathbf{x}_t)-\mathbf{y} \right \|^{2} +\lambda \sum_{l=1}^{D}\left\|\mathbf{w}_p^{l} \right\|^{2} \right)
\end{equation}
The parameter $\beta_t \geq 0$ is the impact of sample $\mathbf{x}_t$.

At the same time, The closed-form solution in Eq. (\ref{eq:wsolution}) can be extend to time series. 
\begin{equation}
\label{eq:rnnsolution}
 \hat{\mathbf{w}}_p^l = \frac{\sum_{t=1}^{p} \beta _t \hat{\mathbf{y}}^* \odot \hat{\varphi}^{l}(\mathbf{x}_t) }{\sum_{t=1}^p \beta _t \left( \sum_{k=1}^{D} \hat{\varphi}^k({\mathbf{x}}_t) \odot  (\hat{\varphi}^k({\mathbf{x}_t}))^*+\lambda \right) }
\end{equation}

The advantage of this incremental update is that we don't have to maintain a large sample set and only need small footprint. In addition, the DCFNet in the online tracking process can be regarded as a RNN network as shown in Fig. \ref{fig:rnn}.
\begin{figure}[htb]
\centering
\includegraphics[trim=0.5cm 6cm 0.4cm 5cm,clip,width=8.5cm]{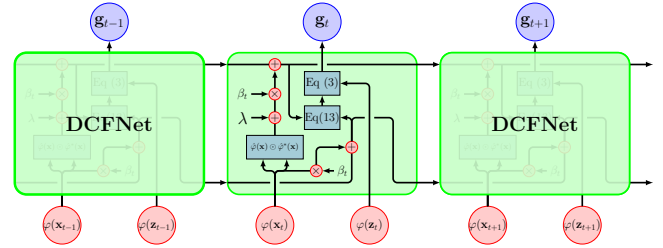}
\vspace{-0.2cm}
\caption{The online tracking process of DCFNet. The Numerator and Denominator of $\hat{\mathbf{w}}_p$ are recurrently forward-propagated and updated as Eq. (\ref{eq:rnnsolution}).}
\label{fig:rnn}
\end{figure}

\vspace{-0.4cm}
\section{EXPERIMENTS}
\vspace{-0.3cm}
\label{sec:experiment}
In this section, we perform an in-depth analysis of the network architectures on OTB \cite{OTB2013,OTB100} and VOT2015 \cite{VOT2015}, the results demonstrate that the end-to-end learning can improve the performance by a significant margin and our DCFNet can get a great balance between accuracy and speed.
\vspace{-0.4cm}
\subsection{Implementation details}
\vspace{-0.1cm}
The convolutional layers of our lightweight network (only 75KB) consist of conv1 from VGG \cite{VGG} with all pooling layers removed and the output forced to 32 channels. Our training videos come from NUS-PRO \cite{NUSPRO}, TempleColor128 \cite{TC128} and UAV123 \cite{UAV123} excluding the videos that overlap with the test set, leading to a total of 166,643 frames. For each video, we choose each pair of frames within the nearest 10 frames, and fed the cropped pair of target patches of $1.5\times$ padding size to the network, resulting 1,651,360 pairs in total. The cropped inputs are resized to $125\times 125$. We apply stochastic gradient descent (SGD) with momentum of 0.9 to train the network from the scratch and set the weight decay $\gamma$ to 0.0005, the learning rate is set to 1e-5. The model is trained for 20 epoch with a mini-batch size of 16.

For the hyper-parameter in the correlation filter layer, we fix the online learning rate $\beta_t$ to 0.008. The regularization coefficient $\lambda$ is set to 1e-4 and the gaussian spatial bandwidth is set to 0.1 for both online tracking and offline training. Similar to \cite{JSSC}, we use patch pyramid with the scale factors $\left\{ a^s|a=1.0375,s=\lfloor-\frac{S-1}{2}\rfloor,\lfloor-\frac{S-3}{2}\rfloor,...,\lfloor\frac{S-1}{2}\rfloor \right\}$.

The proposed DCFNet is implemented in MATLAB with MatConvNet \cite{MatConvNet}. All experiments are conducted on a workstation with Intel Xeon 2630 at 2.4GHz and a single NVIDIA GeForce GTX 1080 GPU. The code is available at:\href{https://github.com/foolwood/DCFNet}{\textit{https://github.com/foolwood/DCFNet}}.
\vspace{-0.5cm}
\subsection{Experiments analyses}
\vspace{-0.2cm}
In this section, we first perform an ablation analysis in terms of network architectures and the number of scale levels impacts on our DCFNet. Then we compare our DCFNet with other correlation filters based trackers and several state-of-the-art CNN based trackers.\\
\vspace{-0.3cm}
\begin{figure}[htb]
\begin{minipage}[b]{.48\linewidth}
  \centering
  \centerline{\includegraphics[trim=2cm 6.8cm 2.5cm 7cm,clip,width=4.2cm]{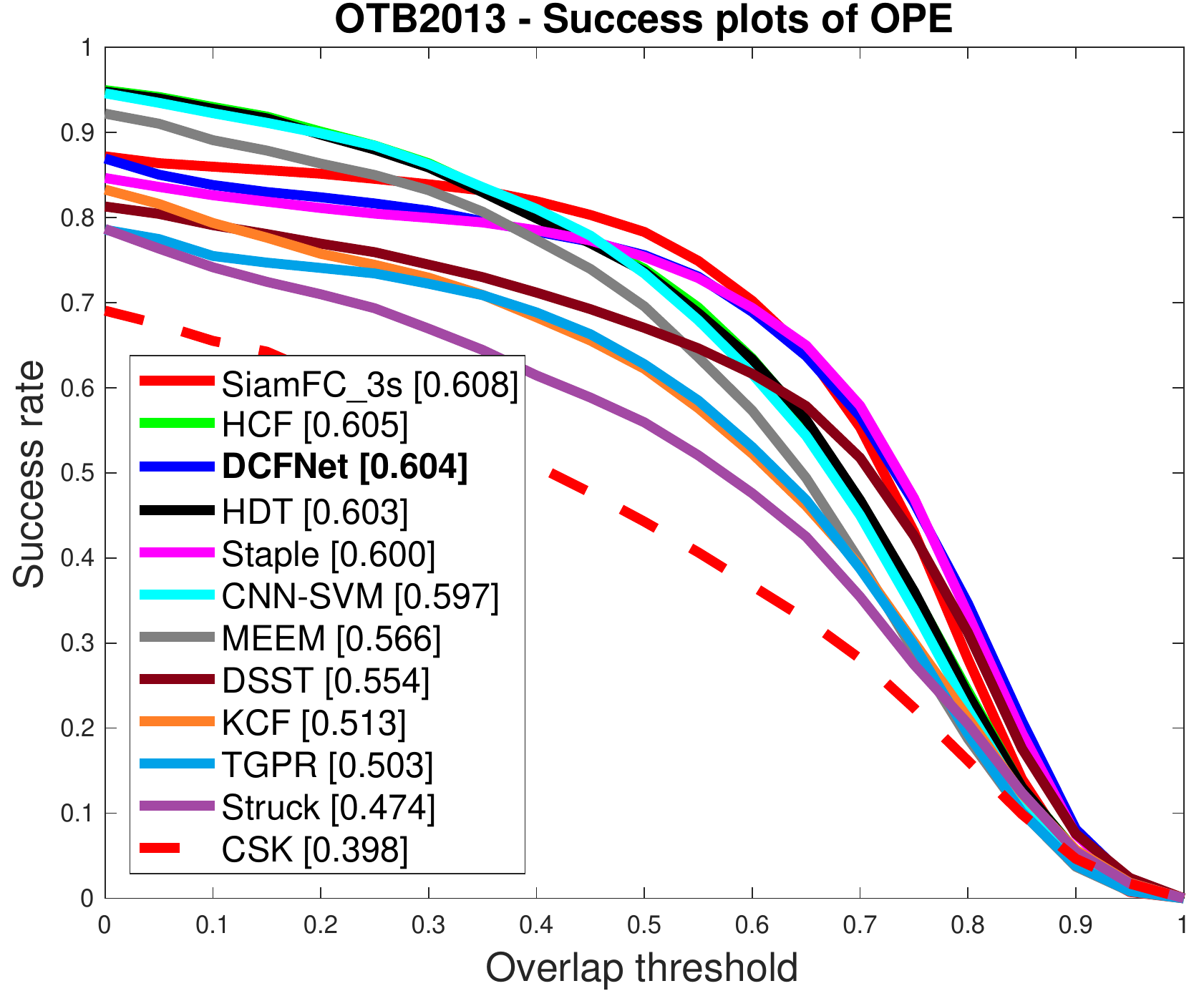}}
  \vspace{-0.2cm}
  \subcaption{\footnotesize{OPE on OTB2013}}
  \label{fig:sub:OPEonOTB2013}
\end{minipage}
\hfill
\begin{minipage}[b]{0.48\linewidth}
  \centering
  \centerline{\includegraphics[trim=2cm 6.8cm 2.5cm 7cm,clip,width=4.2cm]{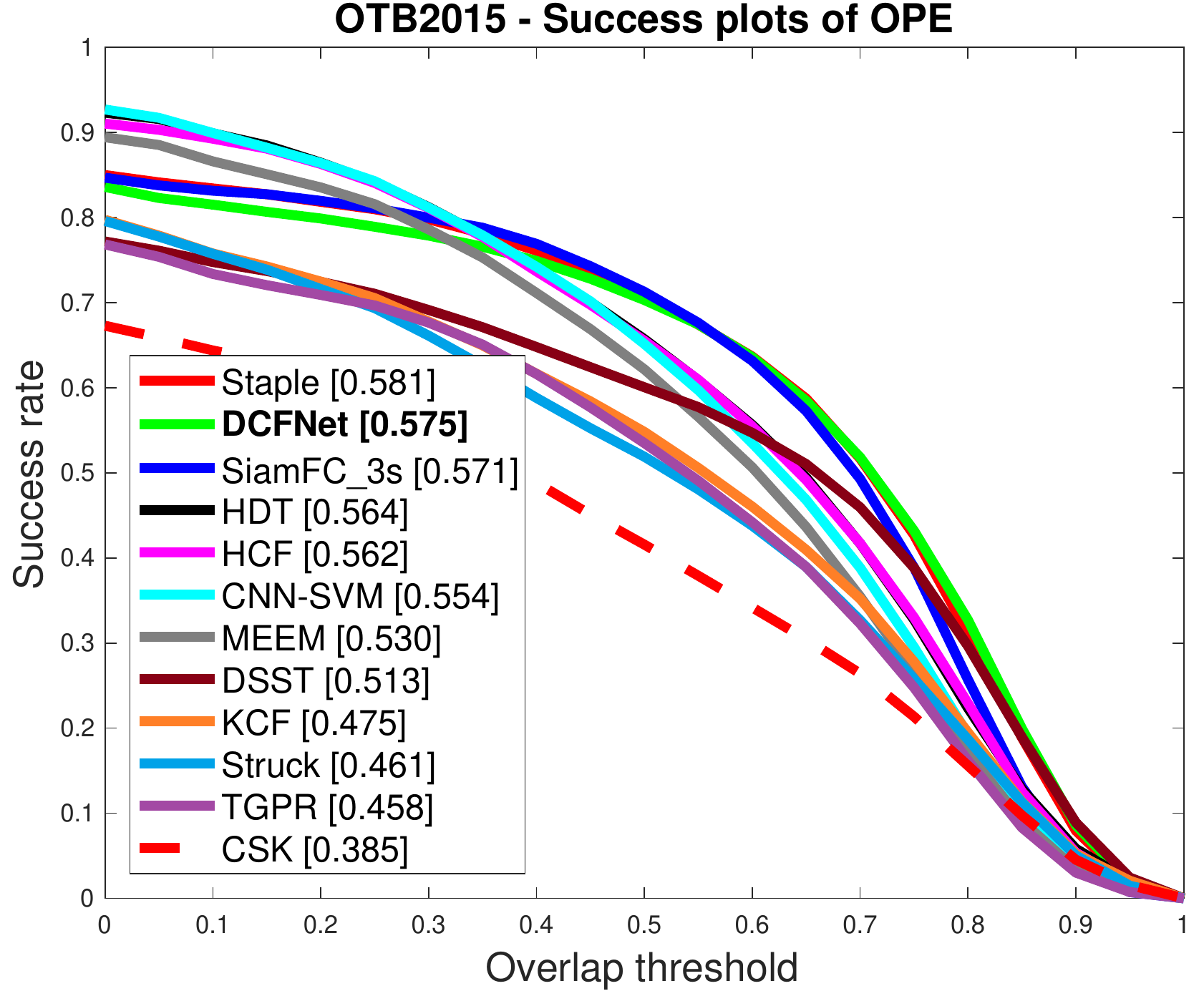}}
  \vspace{-0.2cm}  
  \subcaption{\footnotesize{OPE on OTB2015}}
  \label{fig:sub:OPEonOTB2015}
\end{minipage}
\begin{minipage}[b]{.48\linewidth}
  \centering
  \vspace{0.5cm}
  \centerline{\includegraphics[trim=2cm 6.8cm 2.5cm 7cm,clip,width=4.2cm]{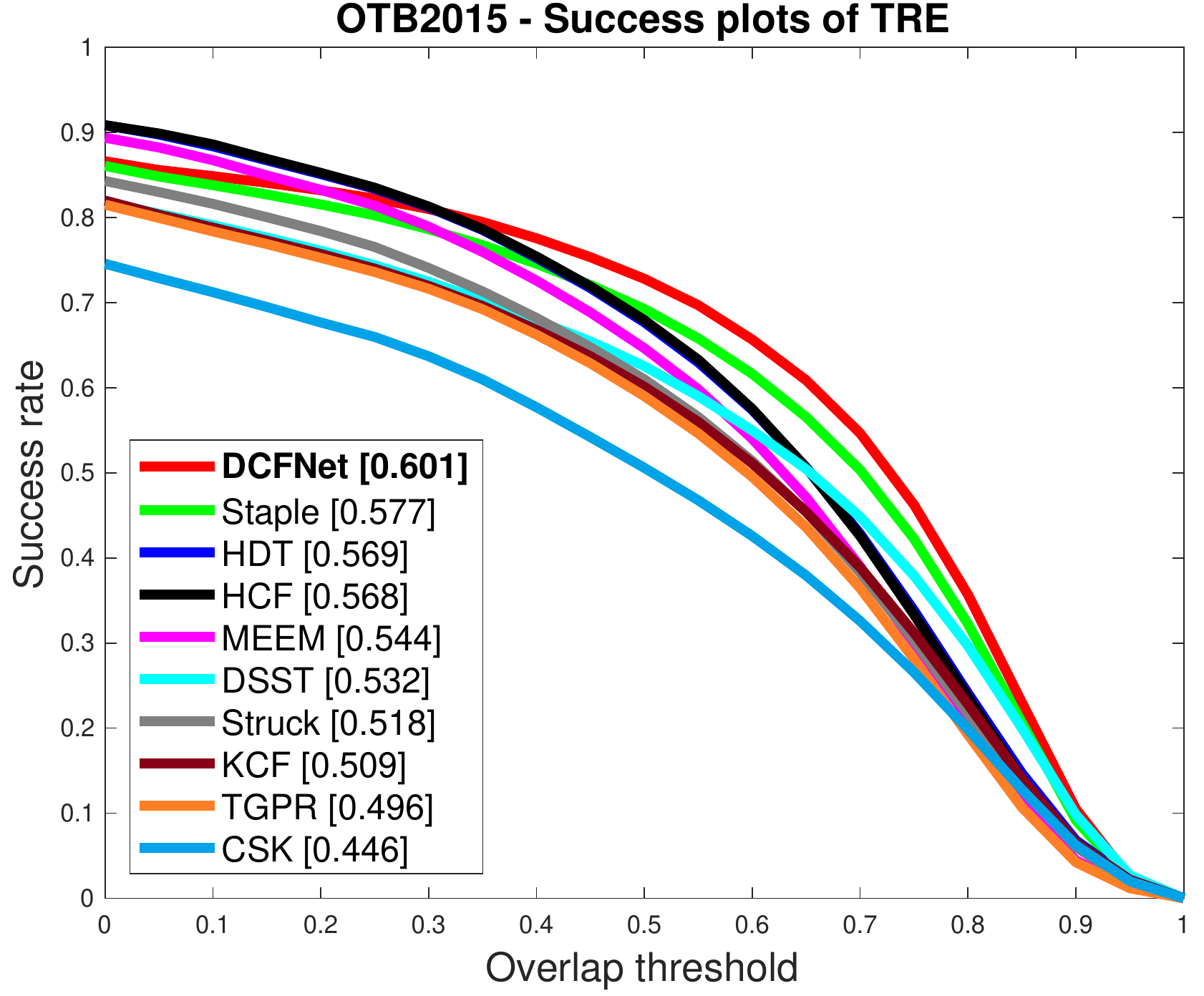}}
  \vspace{-0.2cm}
  \subcaption{\footnotesize{TRE on OTB2015}}
  \label{fig:sub:TREonOTB2015}
\end{minipage}
\hfill
\begin{minipage}[b]{0.48\linewidth}
  \centering
  \vspace{0.5cm}
  \centerline{\includegraphics[trim=2cm 6.8cm 2.5cm 7cm,clip,width=4.2cm]{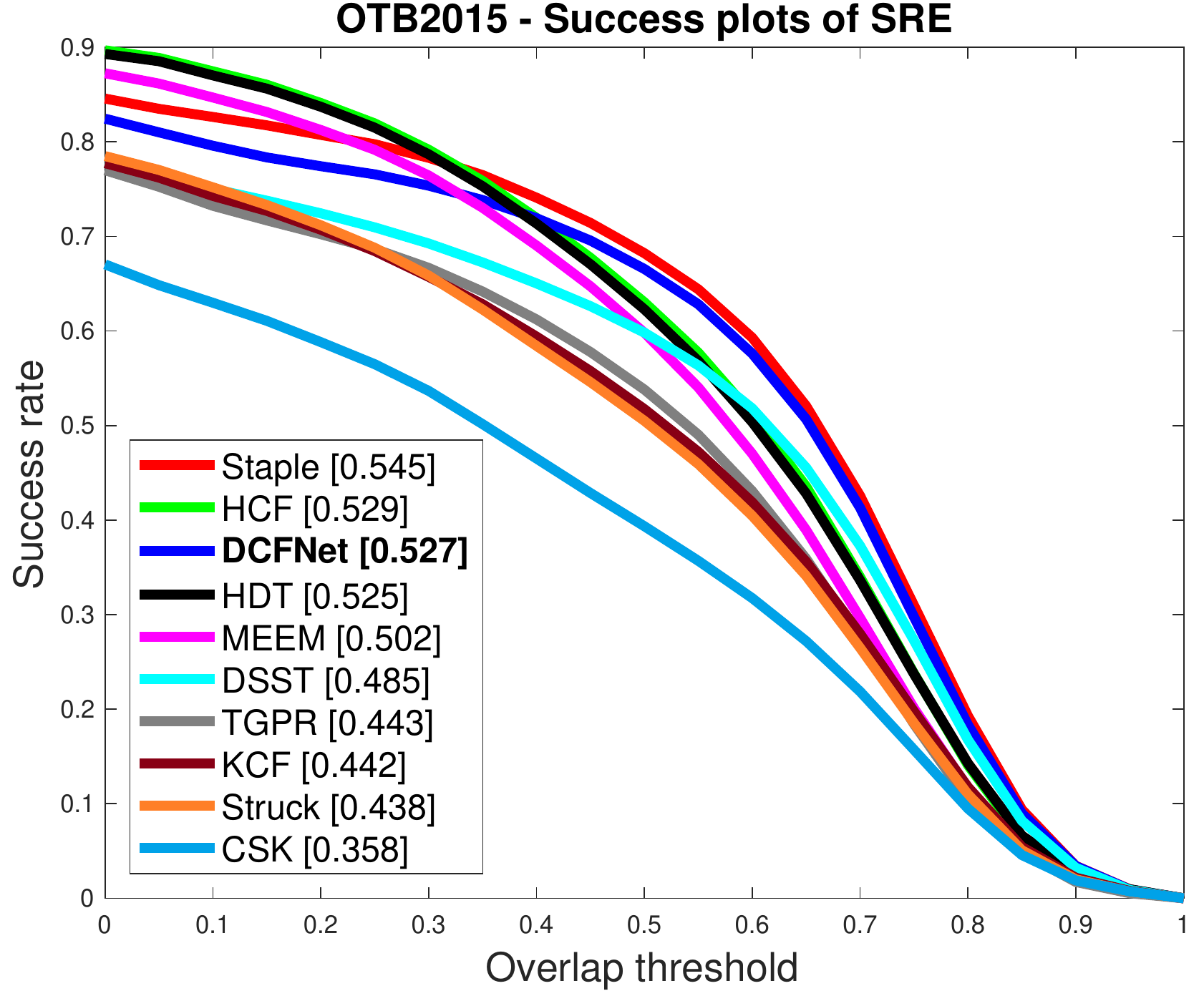}}
  \vspace{-0.2cm}
  \subcaption{\footnotesize{SRE on OTB2015}}
  \label{fig:sub:SREonOTB2015}
\end{minipage}
\vspace{0.2cm}
\caption{Success plots for the OTB2013 \cite{OTB2013}, OTB2015 \cite{OTB100} compared with correlation filters based trackers: CSK \cite{CSK}, KCF \cite{KCF}, DSST \cite{DSST}, HCF \cite{CF2}, HDT \cite{HDT}and others: TGPR \cite{TGPR}, MEEM \cite{MEEM}, Staple \cite{Staple}, CNN-SVM \cite{CNN-SVM}, SiamFC \cite{SiameseFC}.}
\label{fig:resultOTB}
\end{figure}
\begin{table}[htb]
\footnotesize
\centering
\vspace{0.6cm}
\caption{Configurations of ablation study}
\vspace{-0.3cm}
\label{table:configurations}
\begin{tabular}{|c|c|c|c|}
\hline
configurations             & \begin{tabular}[c]{@{}c@{}}DCFNet\\ -conv1\end{tabular} & \begin{tabular}[c]{@{}c@{}}DCFNet\\ -conv1-dilation\end{tabular} & \begin{tabular}[c]{@{}c@{}}DCFNet\\ -conv2\end{tabular} \\ \hline
input                      & \multicolumn{3}{c|}{$125 \times 125$ RGB image}                                                                                                                                             \\ \hline
conv-3-64-dilation-1-relu  & $\surd$                                                 & $\surd$                                                          & $\surd$                                                 \\ \hline
conv-3-32-dilation-1       & $\surd$                                                 &                                                                  &                                                         \\ \hline
conv-3-32-dilation-2       &                                                         & $\surd$                                                          &                                                         \\ \hline
conv-3-64-dilation-1-relu  &                                                         &                                                                  & $\surd$                                                 \\ \hline
conv-3-128-dilation-1-relu &                                                         &                                                                  & $\surd$                                                 \\ \hline
conv-3-32-dilation-1       &                                                         &                                                                  & $\surd$                                                 \\ \hline
LRN                        & $\surd$                                                 & $\surd$                                                          & $\surd$                                                 \\ \hline
\end{tabular}
\vspace{-0.6cm}
\end{table}\\
\textbf{Dataset.} The OTB \cite{OTB2013,OTB100} is the standard benchmark for visual tracking which contains 100 fully annotated targets with 11 different attributes. We follow the protocol of OTB and report the results based on success plots and precision plots for evaluation. The success plots show the percentage of frames in which the overlap score exceeds a threshold; the precision plots show the percentage of frames where the center location error is within a threshold. The VOT challenge \cite{VOT2015} is one of the most influential and largest annual events in tracking field. In the VOT2015 \cite{VOT2015}, a new measure called the expected average overlap (EAO) is introduced which can quantitatively analyze the performance for short-term tracking.\\
\textbf{Ablation study.} We conduct the ablation study on OTB2013. In terms of the network architectures, it proves that the number of training parameters and the receptive field gradually increase with the convolutional layers going deeper. From Table \ref{table:ablation} we see that, our DCFNet with only conv1 achieves better performance compared with deeper conv2, which may not adhere to our intuition. It is probably because that our training set with only 274 objects is not enough for training deeper conv2 from the scratch. To give a better insight into this observation, we modify our DCFNet with conv1 using dilation convolution to approximate the receptive field of deeper conv2. This new variant with a small quantity of parameters also performs better than deeper conv2, and even better than the original conv1 under the CLE metric.
\begin{figure}[htb]
\begin{minipage}[b]{1.0\linewidth}
\centering
\centerline{\includegraphics[trim=2.8cm 11.5cm 2.8cm 11.8cm,clip,width=8.5cm]{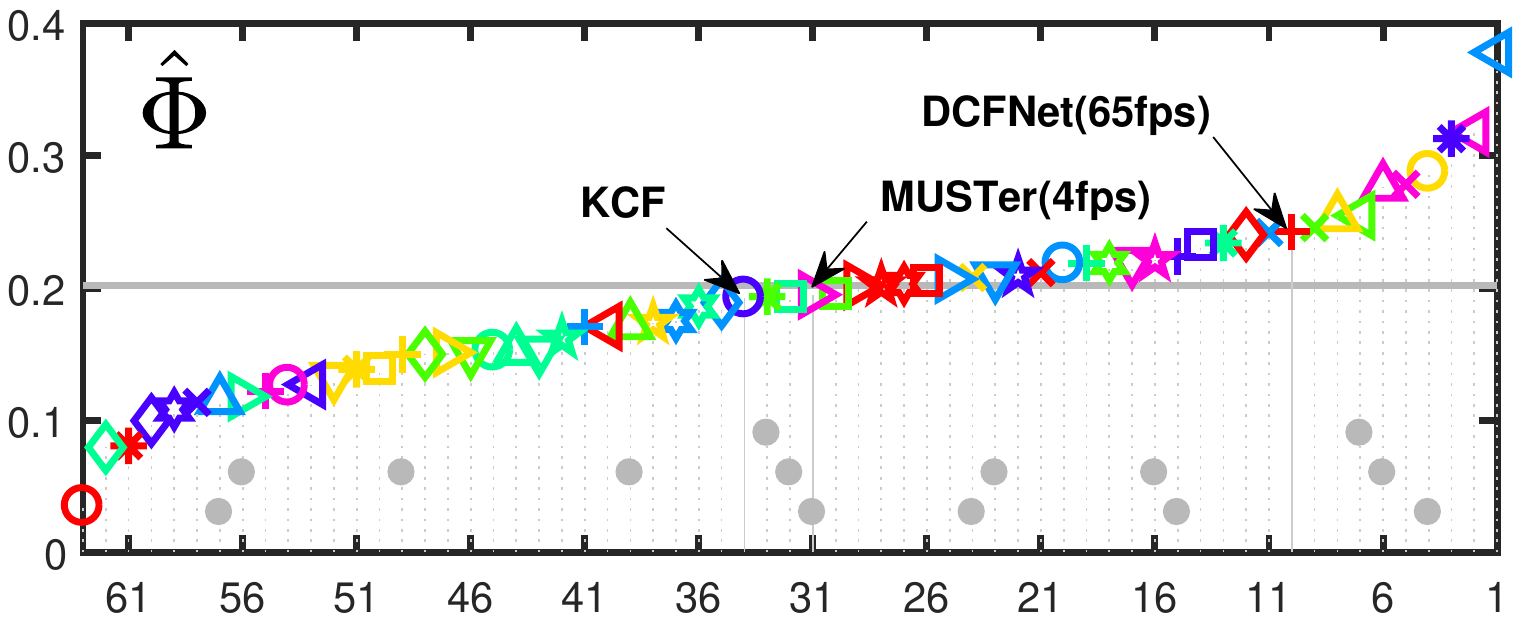}}
\end{minipage}
\vspace{-0.6cm}
\caption{Expercted average overlap plot on VOT2015 \cite{VOT2015}, the full legends are consistent with VOT2015 \cite{VOT2015}}
\label{fig:resultVOT}
\end{figure}
\begin{table}[htb]
\footnotesize
\vspace{0.8cm}
\caption{Ablation study of DCFNet with different architecture and different number of scale levels on OTB2013 \cite{OTB2013} using mean overlap precision (OP) at the threshold of 0.5, mean distance precision (DP) at 20 pixels, mean center location error (CLE) and mean speed.}
 \vspace{-0.2cm}
\begin{tabular}{ c c c c c }
\hline
Tracker & OP & DP & CLE & FPS\\
\hline
DCFNet-conv1 & \textbf{0.89} & 0.88 & 11.52 & 65.94 \\
DCFNet-conv2 & 0.84 & 0.83 & 14.38 & 43.26  \\
DCFNet-conv1-dilation & 0.87 & 0.88 & 10.98 & 63.46 \\
DCFNet-conv1-1s & 0.73 & 0.80 & 18.32 & \textbf{89.44 } \\
DCFNet-conv1-5s & 0.86 & 0.83 & 14.48 & 41.28 \\
DCFNet-conv1-7s & 0.88 & \textbf{0.90} & \textbf{10.18} & 36.86 \\
 \hline
\end{tabular}
\centering
\label{table:ablation}
\vspace{-0.5cm}
\end{table}

In terms of the number of scale levels, we test another 3 settings ($S = 1, 5, 7$), and find that 3-layers design is a great balance between performance and tracking speed.\\
\textbf{Comparison on OTB.} Fig. \ref{fig:resultOTB} shows the results of DCFNet on OTB2013 and OTB2015. Our simple feature training leads to $\mathbf{10\%}$ and $\mathbf{6.2\%}$ gains in success plots on OTB2015 compared with KCF and DSST using HOG features respectively. Although our tracking network is shallower than \cite{CF2, HDT, CNN-SVM}, we can achieve competitive performance with much faster speed. With a more robust online update strategy, the proposed method works slightly better compared to the recent SiamFC \cite{SiameseFC}. Besides the one-pass evaluation (OPE), the temporal robustness evaluation (TRE) and the spatial robustness evaluation (SRE) are conducted to verify the robustness of our DCFNet. In the Fig. \ref{fig:sub:TREonOTB2015}, our DCFNet tracker can achieve a significant improvement in TRE.\\
\textbf{Comparison on VOT.} In the Fig. \ref{fig:resultVOT}, the horizontal dashed line is the VOT2015 state-of-the-art bound. Compared against KCF \cite{KCF} and MUSTer \cite{MUSTer}, we can find our DCFNet can achieve a great balance between performance and speed.
\vspace{-0.3cm}
\section{CONCLUSION}
\vspace{-0.3cm}
\label{sec:conclusion}
We have demonstrated how we can train a lightweight network in an end-to-end fashion to automatically learn the features best fitting DCF based tracking. The feature extraction in the common DCF based trackers can be substituted by using our trained convolutional features, allowing our tracker to achieve a significant accuracy gain compared with these using HoGs. Evaluations on several benchmarks also demonstrates our tracker is more compact and much faster than several state-of-the-art deep learning based trackers. Deep architecture may bring more robust feature extractor while using larger training sets, which is left as our future work.



\footnotesize
\bibliographystyle{IEEEbib}
\bibliography{strings,refs}

\end{document}